\documentclass{article}
\usepackage{amssymb,amsmath,epsfig,epstopdf,color,graphicx,subfigure}
\usepackage{epstopdf,array}
\newcolumntype{L}{>{\centering\arraybackslash}m{3cm}}
\usepackage{arxiv}

\usepackage[utf8]{inputenc} 
\usepackage[T1]{fontenc}    
\usepackage{hyperref}       
\usepackage{url}            
\usepackage{booktabs}       
\usepackage{amsfonts}       
\usepackage{nicefrac}       
\usepackage{microtype}      

\title{Winning with Simple Learning Models: Detecting Earthquakes in Groningen, the Netherlands}

\author{
  Umair bin Waheed\\
  Department of Geosciences\\
  King Fahd University of Petroleum and Minerals
  \And
  Ahmed Shaheen\\
  Department of Geosciences\\
  King Fahd University of Petroleum and Minerals
   \And
 Mike Fehler\\
  Earth Resources Laboratory\\
  Massachusetts Institute of Technology
    \And
  Ben Fulcher\\
  School of Physics\\
  The University of Sydney
}

\begin{document}
\maketitle

\begin{abstract}
Deep learning is fast emerging as a potential disruptive tool to tackle longstanding research problems across the sciences. Notwithstanding its success across disciplines, the recent trend of the overuse of deep learning is concerning to many machine learning practitioners. Recently, seismologists have also demonstrated the efficacy of deep learning algorithms in detecting low magnitude earthquakes. Here, we revisit the problem of seismic event detection but using a logistic regression model with feature extraction. We select well-discriminating features from a huge database of time-series operations collected from interdisciplinary time-series analysis methods. Using a simple learning model with only five trainable parameters, we detect several low-magnitude induced earthquakes from the Groningen gas field that are not present in the catalog. We note that the added advantage of simpler models is that the selected features add to our understanding of the noise and event classes present in the dataset. Since simpler models are easy to maintain, debug, understand, and train, through this study we underscore that it might be a dangerous pursuit to use deep learning without carefully weighing simpler alternatives.
\end{abstract}


\section{Introduction}
Deep learning is gaining widespread attention due to its success on applications across scientific disciplines. This is mainly driven by its ability to find complex patterns in large datasets without the need for feature extraction or engineering. It comes as no surprise that seismologists, like domain experts from other scientific disciplines, are starting to find value in deep learning. Numerous articles demonstrating efficacy of deep learning to longstanding seismological problems have appeared in the recent past. These include problems such as earthquake detection, ground-motion prediction, seismic tomography, and earthquake early warning among many others~\cite{kong2019machine}.

However, there is a growing concern among some practitioners on the overuse of deep learning in the sciences. For example, \cite{mignan2019one} showed that a logistic regression model with only two trainable parameters achieves similar or better performance than that of a 13,451-parameter deep neural network of~\cite{devries2018deep} for the problem of aftershock prediction. \cite{dittgen2019empirical} also demonstrated, on a number of classification problems, that simple models with feature extraction perform as well as complex, uninterpretable black box models. 


Here we study the problem of detecting earthquakes in the Groningen area using a simple machine learning model. While complex algorithms such as convolutional neural networks have been shown to detect buried events by feeding in raw seismograms~\cite{perol2018convolutional}, we are interested in exploring whether low magnitude events can be detected with a simple learning model such as logistic regression. To do so, however, we need well-discriminating features between waveforms caused by an earthquake versus other sources (noise). To this end we use {\it hctsa} -- a software tool for massive feature extraction from time-series data~\cite{fulcher2017hctsa}. This collection of more than 7700 time-series features allows us interdisciplinary breadth to extract meaningful features for the problem at hand. We train the model on already cataloged events using selected well-discriminating features and then test the model on a continuous seismic record. We detect five new events in a four-hour continuous seismic record, in addition to two events already present in the catalog.

\section{Methodology}

\subsection{Data selection and pre-processing}

To train our model, we use earthquake data from the Groningen gas field located in the north-east part of the Netherlands. Groningen is the largest gas field in Western Europe, producing natural gas since 1963~\cite{willacy2018application}. The area has witnessed a gradual increase in seismic activity related to reservoir compaction over the years with M > 3 earthquakes occurring in 2013, 2018, and recently in May, 2019. 

The recurring occurrence of these events encouraged the expansion of seismic network over the years to study these earthquakes and their relationship with reservoir production rates~\cite{van2015induced}. Data from these networks is made publicly available by the Royal Netherlands Meteorological Institute (KNMI). We use 3C seismic data recorded on the G-network, which consists of 70 shallow boreholes with four velocity sensors deployed at 50~m depth intervals. We use data from 47 events identified in the KNMI catalog occurring between October 1, 2017 and February 28, 2018. These events range from M0.2 to M3.4 with the majority of earthquakes between M0.5 - M1.0. 

We trim a 20~s window around each recorded event before pre-processing. The pre-processing steps include demeaning, detrending, bandpass filtering (5 - 25 Hz), and amplitude normalization to unity. This set constitutes our event class. Next, we extract noise windows from waveform records outside of the identified event times and run them through the same pre-processing steps to have our noise class. We manually verify all windows to ensure integrity of the two data classes. We end up with about 2300 windows as event labels and about 4000 noise labels. Feature extraction is performed on this labeled dataset before using them to train the model.

\subsection{Feature analysis and selection}

The {\it hctsa} package lets us compute over 7700 time-series features on these windows using operations such as basic statistics of the distribution, linear correlation, stationarity, information theoretic and entropy measures, linear and nonlinear model fits, wavelet methods, etc. This gives us a rich set of features to work with compared to previous studies looking into the feature extraction problem for earthquake detection using hand-selected features~\cite{akram2017robust,qu2018automatic}.

On the windowed dataset, we obtain $\sim$ 6800 well-behaved features. We filter out features that are not applicable to the dataset. This is usually signaled through features with special values such as {\it Nan} or {\it inf}. To meaningfully compare the features with different ranges and distribution of output, an outlier-robust sigmoidal normalizing transformation is applied to the well-behaved features. Using the event and noise labels in the data, we rank the selected features based on their accuracy in differentiating between the two classes individually using a linear classifier. The top 50 features are then compared using pairwise correlation coefficients for the given dataset to select the most uncorrelated features in discriminating the two classes. As a result, we end up with only four well-discriminating features. Table~\ref{table1} lists the operation names from the {\it hctsa} package resulting in these four features and their description. While it is possible to introduce self-designed features, we rely on the existing ones only as the list of features in the {\it hctsa} package is already quite exhaustive.

\begin{table}[ht]
\begin{center}
    \begin{tabular}{|c|c|}
      \hline
      {\bf Feature name} & {\bf Description} \\
      \hline
      \hline
      DN\_RemovePoints\_min\_05\_fzcacrat & \multicolumn{1}{m{8cm}|}{Measures how time-series properties change as points are removed. Specifically, it computes the first zero-crossing of the normal linear autocorrelation function as 50\% of the lowest values are removed.} \\
      \hline
      SY\_SlidingWindow\_s\_s\_5\_1 & \multicolumn{1}{m{8cm}|}{Sliding window measure of stationarity. Specifically, it divides the time-series into five windows and computes the standard deviation in each window followed by computing standard deviation of the resulting five standard deviation values.}  \\
      \hline
      ST\_MomentCorr\_002\_02\_mean\_std\_sqrt\_mi &\multicolumn{1}{m{8cm}|}{Measures correlations between simple statistics in local windows of a time-series. Specifically, it computes mutual information between two vectors formed by computing the mean and standard deviation of the time-series in a sliding window. The length of the sliding window is 2\% of the entire time-series with a 20\% overlap between consecutive windows.} \\
      \hline
      FC\_Surprise\_dist\_100\_5\_q\_500\_tstat & \multicolumn{1}{m{8cm}|}{Measures the level of surprise due to the next data point given recent memory. Specifically, it coarse-grains the time-series into five groups and computes a summary of information gain with 100 previous memory samples. }   \\
      \hline
    \end{tabular}
\end{center}
\caption{List of selected four features and their description.}
\label{table1}
\end{table}

\section{Results}

Having selected the top discriminating features, we do a 70/30 split of the windowed data into training and test sets. We train a logistic regression model using features computed from the training data windows.  We obtain a training accuracy of 99.27\%. Then we apply the trained model on the other 30\% of the windowed data for a preliminary test. The accuracy of the trained model on the test set is found to be 99.34\%. With only five trainable parameters for the logistic regression model (four weights for each input feature and a bias term), we observe quite high training and test accuracy. This is mainly due to the selected features, which also helps add to our understanding about the two classes in our data.

Next we apply the trained model on a four-hour continuous record starting from 00:00:00 on November 1, 2016. Two events of M1.9 and M2.2 have already been identified during this period in the KNMI catalog. The hypocenter locations for the two events were estimated to be quite close to each other. Therefore, by applying the trained model we hope to detect the two identified events plus some other buried events missing in the catalog.

We consider data from five stations positioned around the epicenter of the two identified events (namely G19, G23, G24, G29, G67). Specifically, we use data from the sensor at the deepest level in the shallow boreholes (200~m deep). We partition each seismogram into 20~s windows and apply the same pre-processing routine as the training data. Next, we compute the four features listed in Table~\ref{table1} for each 20~s window. We input these feature values to the trained logistic regression model and obtain the probability of an earthquake occurring within a 20~s window interval for each sensor separately. To minimize false alarms, we require a minimum of two sensors to flag an earthquake within the same 20~s window. 

\begin{figure}[ht!]
\begin{center}
\includegraphics[width=1\textwidth]{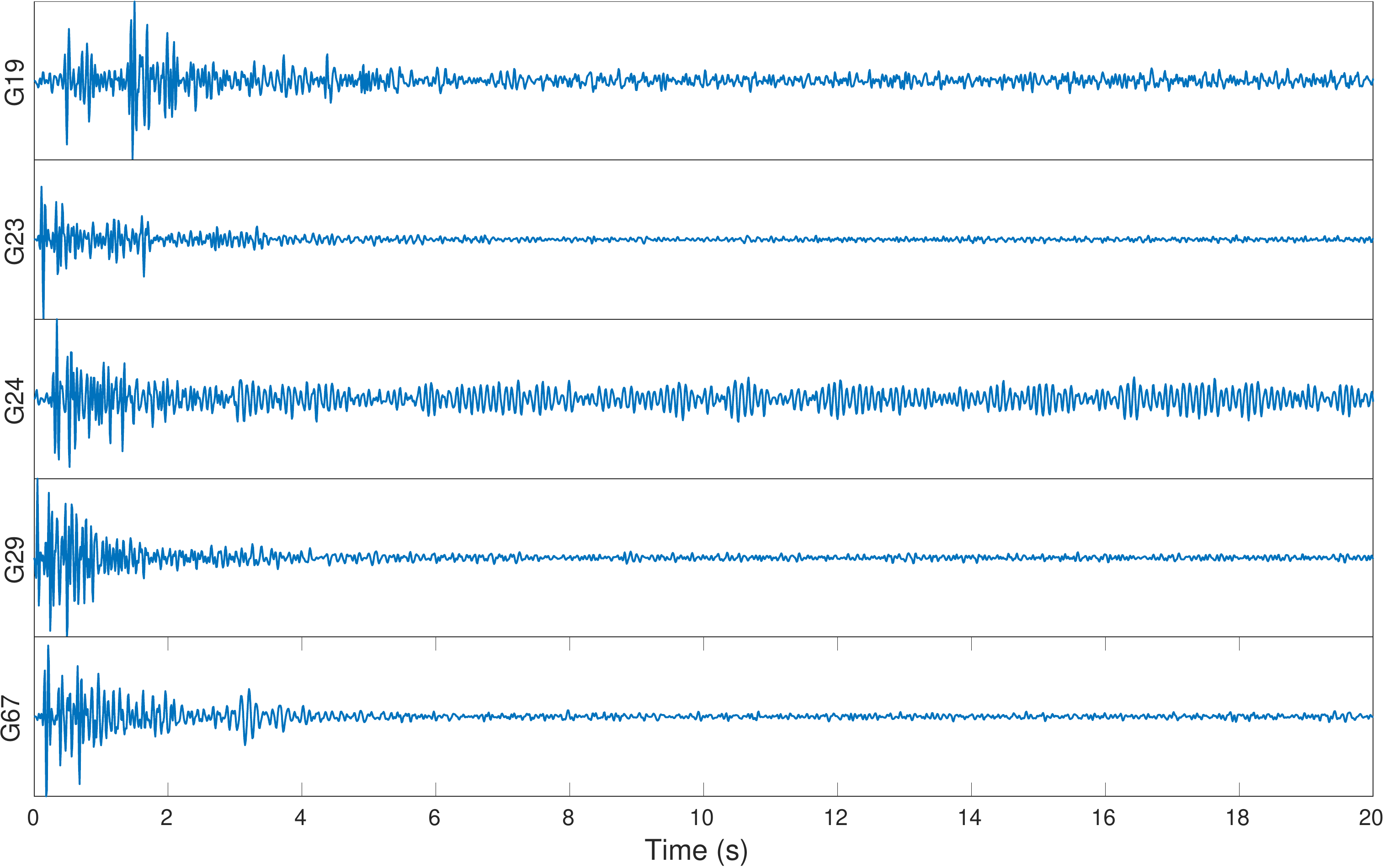}
\end{center}
\caption{
Processed seismogram windows recorded on the vertical component of the five sensors used for continuous data test. The start of the window is at 02:03:20 on November 1, 2016. The event was detected on all five sensors using the trained logistic regression model.
}%
\label{fig1}
\end{figure}

Table~\ref{table2} lists the earthquakes detected by our trained logistic regression model. The table lists the start time of the 20~s window containing recorded earthquake energy. It also lists the number of sensors the event was detected on. It is no surprise that the two already cataloged earthquakes were detected on all five sensors. Other detected events, that are not in the catalog, were manually verified to be true events. Figure~\ref{fig1} shows seismograms recorded on the vertical component of the five sensors for a detected event that is not present in the KNMI catalog. 

Although the model also detects several events on a single sensor, we discard those to minimize the probability of producing false flags. Through template matching and visual inspection, we find that while most of those events were indeed false flags, we also miss a couple of buried events that were detected only on a single sensor. 

\begin{table}[ht]
\begin{center}
    \begin{tabular}{|c|c|}
      \hline
      {\bf Event window start time} & {\bf Number of sensors} \\
      \hline
      \hline
      {\bf 00:12:20} & {\bf 5} \\
      \hline
      00:12:40 & 5  \\
      \hline
      00:16:20 & 2 \\
      \hline
      00:32:40 & 2 \\
      \hline
      00:56:00 & 3 \\
      \hline
      {\bf 00:57:40} & {\bf 5} \\
      \hline
      02:03:20 & 5 \\
      \hline
    \end{tabular}
\end{center}
\caption{Event window start times and the number of sensors that detected the event for the four-hour continuous data recorded on November 1, 2016. The two relatively larger earthquakes, already present in the KNMI catalog, are highlighted in bold.}
\label{table2}
\end{table}

\section{Conclusions}
\vspace{-0.2cm}
In a diversion from the norm of using complex learning models, we explored the possibility of using a logistic regression model with feature extraction to detect induced earthquakes in Groningen, the Netherlands. We used the {\it hctsa} package for massive feature extraction from the recorded seismograms. Starting with over 7700 features collected from across scientific disciplines, we found four well-discriminating features for the event and noise classes. On a continuous four-hour record, the trained network identified five low-magnitude events, in addition to the two already present in the KNMI catalog. 

The real value in simple learning models lies in their ease of understanding, debugging, and training. Moreover, they have better interpretability compared to complex models that are often used as black boxes. The identified features also add to our understanding about the different classes present in the data. As a starting point, we used a logistic regression model to automatically detect induced seismic events with feature extraction. A detailed study on a diverse dataset with performance comparisons against complex models is needed to further our understanding in this regard. Moreover, the proposed approach can be used to explore other problems in Seismology to address issues such as opaqueness and overfitting often associated with black-box models, without compromising on performance.

\bibliographystyle{unsrt}  
\bibliography{references}

\end{document}